\documentclass[sigplan,10pt,nonacm]{acmart}
\renewcommand\footnotetextcopyrightpermission[1]{}
\usepackage{booktabs}
\usepackage{algorithm}
\usepackage{algorithmicx}
\usepackage[noend]{algpseudocode}
\usepackage{graphicx}
\usepackage{subfigure}
\usepackage{multirow}
\usepackage{enumitem}
\usepackage{float}
\usepackage{placeins}

\makeatletter \newcommand{\removelatexerror}{\let\@latex@error\@gobble} \makeatother

\usepackage{xspace}

\newcommand{\ours}{\textsc{SpecLA}\xspace}

\newcommand{\stitle}[1]{\par\noindent\textbf{#1}}     

     \newcommand{\eat}[1]{}

\AtBeginDocument{%
  }

\begin{document}

\title[SpecLA]{\ours: Efficient Speculative Decoding for Linear-Attention Models}

\author{Zhibin Wang}
\affiliation{%
  \department{State Key Laboratory of Novel Software Technology}
  \institution{Nanjing University}
  \city{Nanjing}
  \country{China}}

\author{Xuying Han}
\affiliation{%
  \department{State Key Laboratory of Novel Software Technology}
  \institution{Nanjing University}
  \city{Nanjing}
  \country{China}}

\author{Zhaohua Yang}
\affiliation{%
  \department{State Key Laboratory of Novel Software Technology}
  \institution{Nanjing University}
  \city{Nanjing}
  \country{China}}

\author{Fuliang Liu}
\affiliation{%
  \department{State Key Laboratory of Novel Software Technology}
  \institution{Nanjing University}
  \city{Nanjing}
  \country{China}}

\author{Xue Li}
\affiliation{%
  \institution{Alibaba Group}
  \city{Hangzhou}
  \country{China}}

\author{Rong Gu}
\affiliation{%
  \department{State Key Laboratory of Novel Software Technology}
  \institution{Nanjing University}
  \city{Nanjing}
  \country{China}}

\author{Sheng Zhong}
\affiliation{%
  \department{State Key Laboratory of Novel Software Technology}
  \institution{Nanjing University}
  \city{Nanjing}
  \country{China}}

\author{Chen Tian}
\affiliation{%
  \department{State Key Laboratory of Novel Software Technology}
  \institution{Nanjing University}
  \city{Nanjing}
  \country{China}}

\renewcommand{\shortauthors}{Wang et al.}

\begin{abstract}
Linear-attention models replace the growing KV cache with recurrent states, but autoregressive decoding still reads, updates, and writes these states one token at a time. Speculative decoding can reduce this cost by verifying several draft tokens in one target pass, yet existing speculative systems are designed for Transformer KV caches. For stateful linear-attention targets, verification must follow recurrent dependencies across chains and branches, acceptance must update only the accepted state trajectory, and the drafter must avoid submitting candidates that waste stateful verification work. This paper presents \ours, a speculative decoding runtime for stateful linear-attention models. \ours verifies chains and trees with topology-aware kernels, stores compact factors produced during verification to recover accepted states, and uses confidence pruning plus a target-aligned EAGLE-style drafter to feed useful candidates to the verifier. On an NVIDIA H100 with a public GDN-1.3B target, \ours achieves up to 1.70$\times$ end-to-end speedup over autoregressive decoding.
\end{abstract}

\eat{


}

\maketitle

%% ================================================================
%% Sections — each file contains one \section{...} and its content
%% ================================================================

\section{Introduction}

Autoregressive decoding remains a central bottleneck in large language model serving. Each generated token requires a target-model invocation, and for memory-bound decode paths the latency is dominated not only by arithmetic but also by cache access, state movement, synchronization, and kernel launch overheads~\cite{kwon2023pagedattention,qwen2025qwen3}. Speculative decoding attacks this bottleneck by separating generation into a fast draft stage and a target verification stage: a draft model proposes several future tokens, and the target model verifies them together while preserving the target distribution through posterior acceptance~\cite{leviathan2023speculative,chen2023speculative}. This has become a practical acceleration strategy for Transformer serving, with follow-on systems using feature-level drafting, auxiliary draft heads, and tree-structured verification~\cite{li2024eagle,li2024eagle2,li2025eagle3,cai2024medusa,ankner2024hydra,miao2024specinfer,wu2025stree}, because a submitted draft can be represented as a token-indexed extension of the existing KV-cache history~\cite{kwon2023pagedattention}.

At the same time, linear-attention and stateful sequence models are changing the serving substrate. Instead of storing a growing KV cache, these models compress the prefix into recurrent state and update that state with each new token~\cite{gu2023mamba,yang2024gla,dao2024transformers,yang2024deltarule}. Gated DeltaNet (GDN) further combines gating with Delta-rule updates and appears in recent hybrid and open-weight model stacks~\cite{yang2025gdn,qwen2025qwen3nextcard,nvlabs2026gdnrepo}. Efficient kernels for these layers already require specialized prefill and decode implementations~\cite{yang2024fla,qin2024lightning,beck2025tfla,yang2024deltarule}, and the stateful design does not make decode free: each token still reads, updates, and writes a dense recurrent state~\cite{gupta2026persistent}. Thus, linear-attention decode remains a natural target for speculation, because verifying multiple draft tokens in one target-side pass could amortize recurrent-state movement and fixed launch costs across multiple accepted tokens.

Two direct adaptations fail to transfer speculative decoding to stateful linear-attention targets. \emph{1) Decode-kernel replay} preserves the recurrence, but it also preserves the per-token recurrent-state round trip that speculation is meant to amortize; for branching proposals, different branches additionally require different intermediate recurrent states~\cite{yang2024fla,gupta2026persistent}. \emph{2) Prefill-kernel reuse} exposes token parallelism, but speculative windows are too short to amortize long-prefill setup costs, and branching proposals do not follow the regular causal dependency assumed by prefill kernels~\cite{yang2024fla,beck2025tfla,yang2024deltarule,yang2025gdn}.

In this paper, we aim to make speculative decoding effective for stateful linear-attention targets, which requires addressing three coupled challenges. \emph{1) SRAM-resident verification:} the target kernel must keep recurrent state tiles resident for chain-like candidates, preserve ancestor-only dependencies for branching candidates, and avoid prefill-style setup costs on short speculative windows. \emph{2) Dense-state acceptance:} posterior selection must advance only the accepted recurrent-state updates without full-state snapshots, token replay, or an extra state-access pass at every iteration boundary. \emph{3) Target-aligned drafting:} the proposal path must submit useful candidate work by pruning low-confidence paths, aligning draft features with the recurrent target dynamics, and packing retained candidates into the topology representation consumed by the verifier.

We present \ours, a speculative decoding runtime that preserves the external draft--verify--accept loop but replaces KV-cache suffix manipulation with three target-aware mechanisms: topology-aware verification that uses the submitted topology as the target-kernel schedule, accepted-factor state management that logs compact verification factors instead of dense state endpoints, and target-aligned draft integration that treats proposal generation as cost-aware scheduling for the verifier.

\stitle{Topology-aware verification (Section~\ref{sec:kernel}).}
\ours treats the submitted topology as a target-kernel scheduling signal rather than forcing every proposal through decode replay or prefill reuse. Because draft tokens are known before target execution, chain-like candidates can be verified layer by layer while keeping V-tiled recurrent-state slices resident on chip. Branching proposals instead require ancestor-only state flow: a tree-masked factorized GDN path gives the fully parallel endpoint, while a chain-decomposed hybrid path keeps the lower-overhead serial kernel within dependency-respecting chains and parallelizes across ready chains.

\stitle{Accepted-factor state management (Section~\ref{sec:state-management}).}
Because rejected candidates cannot be removed by truncating recurrent state, the runtime separates speculative progress from durable state. The key observation is that verification already produces the compact layer-specific factors needed to reconstruct accepted state updates, so \ours buffers factors instead of full states or token-only logs. After posterior selection, it gathers only the accepted path and delays the state update so the pending factor buffer is applied inside the next verification kernel rather than through a standalone state-update pass.

\stitle{Target-aligned draft integration (Section~\ref{sec:layer-aware-stack}).}
The draft side is adapted to the recurrent target rather than treated as a Transformer plug-in. Since target-side cost depends on the submitted topology, proposal generation also acts as a scheduling problem: confidence-guided pruning removes low-probability paths before verification, and an EAGLE-style drafter trained on recurrent target features makes these pruning scores reflect the target dynamics~\cite{li2024eagle,li2024eagle2}. The retained candidates are then packed as parent-pointer topologies consumed by the target kernels.

\stitle{Evaluation (Section~\ref{sec:evaluation}).}
We implement \ours in PyTorch~\cite{paszke2019pytorch} and Triton~\cite{tillet2019triton} and evaluate it on an NVIDIA H100 GPU with a public GDN-1.3B target~\cite{map2026gdn13b}. End-to-end, \ours achieves 1.42$\times$, 1.70$\times$, and 1.06$\times$ speedup over autoregressive decoding on the mixed suite, GSM8K, and HumanEval. Microbenchmarks show that chain-decomposed hybrid verification improves over root-to-leaf replay by 1.80--7.11$\times$, factor-buffer reuse reduces accepted-state recovery latency by 2.74--4.28$\times$, and delayed state update reduces commit-boundary latency by 1.15--1.44$\times$.

\section{Background and Motivation}
\label{sec:background}

In this section, we first review two technical areas: speculative decoding and linear attention. We then establish why speculative decoding does not transfer directly to stateful linear-attention models through a naive adaptation. Finally, we identify three challenges for efficiently employing speculative decoding with linear attention, from kernel design to state management to runtime-stack integration.

\subsection{Speculative Decoding}

\begin{figure}[t]
  \centering
  \includegraphics[width=\columnwidth,trim=10bp 38bp 0 8bp,clip]{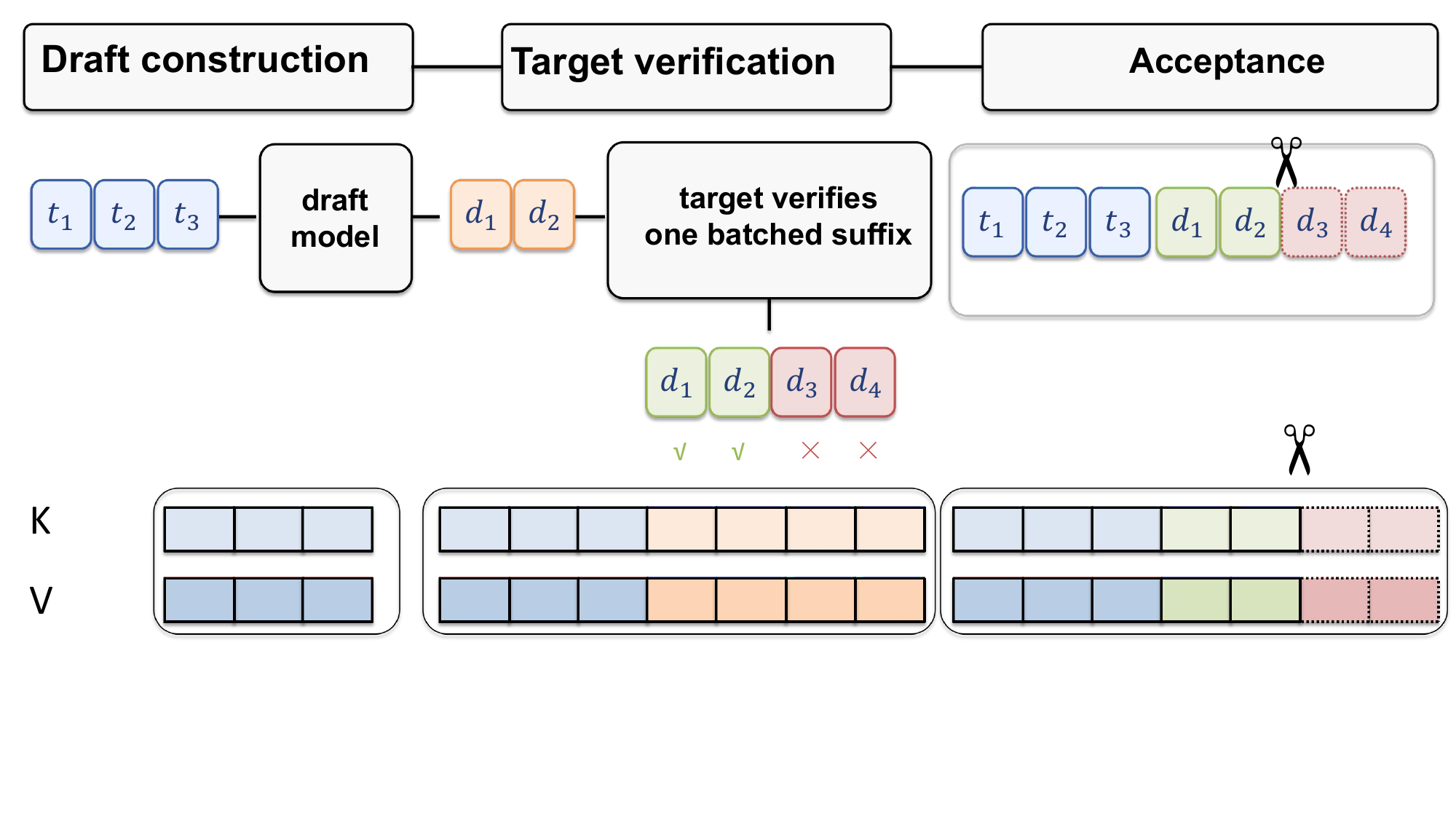}
  \caption{Suffix truncation in Transformer-oriented speculative decoding. Accepted draft tokens remain in the KV cache, while rejected tokens are discarded as a cache suffix.}
  \Description{A prefix stored in the KV cache is extended by four proposed tokens. Target verification accepts the first two proposed tokens and rejects the remaining two. The accepted prefix is kept and the rejected suffix is truncated from the KV cache.}
  \label{fig:sd-workflow}
\end{figure}

Speculative decoding is an inference-time acceleration technique for autoregressive generation: a fast approximation model drafts several future tokens, and the target model verifies them together so that one expensive target invocation can produce multiple accepted tokens while preserving the target model's output distribution through a rejection-sampling-style acceptance rule~\cite{leviathan2023speculative,chen2023speculative}. This is especially useful for memory-bound decoding because target-side cache access, state movement, synchronization, and launch overheads can be paid once for a block of candidates instead of once per generated token. Figure~\ref{fig:sd-workflow} shows the workflow, which we group into three stages throughout this paper. We use speculative decoding under Transformer architectures as the running example.

\stitle{Draft construction.}
The runtime starts from an accepted prefix and constructs speculative work conditioned on that prefix. In the original formulation, the accepted history is represented by token-indexed KV-cache entries, and a faster but less powerful draft model proposes a short token continuation that approximates what the target model will generate~\cite{leviathan2023speculative,chen2023speculative}. Later systems instantiate the same step with feature-level drafting~\cite{li2024eagle,li2024eagle2,li2025eagle3}, auxiliary draft heads~\cite{cai2024medusa,ankner2024hydra}, or tree-structured proposals~\cite{miao2024specinfer,wu2025stree}. This stage fits Transformer runtimes because the draft can be represented as additional token positions, or paths of token positions, extending an already token-indexed history.

\stitle{Target verification.}
The target model scores the submitted candidates in one batched forward pass rather than decoding them one by one, producing the target-side probabilities used by the acceptance rule. This amortizes one expensive target invocation over multiple possible output tokens: parallel scoring of a short continuation can be comparable to a single target decode call, while potentially confirming multiple future tokens~\cite{chen2023speculative}. Tree-based verification extends the same idea by checking multiple candidate continuations in the same target pass~\cite{miao2024specinfer,wu2025stree}. This stage is well matched to Transformers because the committed KV cache can be reused as a shared prefix, while draft-token KV entries form a compact speculative suffix that is produced and accessed in one batched verification pass. For memory-bound Transformer decoding, verification turns repeated per-token KV-cache access into a regular batched access pattern, making cache traffic easier to amortize across accepted tokens.

\stitle{Acceptance and cache commitment.}
Posterior selection applies a modified rejection-sampling rule to keep the accepted prefix of the proposal while preserving the target model's output distribution~\cite{leviathan2023speculative,chen2023speculative}. When a candidate is rejected, generation falls back to a corrected target-side distribution for the next output; the runtime commits the accepted output tokens and removes the rejected suffix from the KV cache. This is cheap because the speculative window is stored as token positions after the committed prefix: rejection is a suffix-truncation operation, not a reconstruction of model state~\cite{kwon2023pagedattention}. This cache-commit step is therefore especially convenient for Transformers, whose generation progress is already organized as a token-indexed cache history.

\subsection{Linear Attention and Stateful Sequence Models}

% 中文概括：linear attention 和 stateful sequence model 用 recurrent state update 代替显式二次 attention，从而让内存不随生成长度增长。
Long-context serving scenarios, including long-document understanding, streaming or multi-turn generation, and retrieval- or code-heavy workloads with large prefixes, put sustained pressure on full-attention models~\cite{kwon2023pagedattention,qwen2025qwen3}. Full attention forms pairwise token interactions, giving prefill and training cost that grows quadratically with sequence length~\cite{vaswani2017attention}; during decoding, the growing KV cache further creates capacity and bandwidth pressure, and per-token cache access remains tied to prefix length~\cite{kwon2023pagedattention}. Linear-attention layers change this execution model: the prefix is compressed into a recurrent state whose size is independent of sequence length, and each new token updates and queries that state through linear-size operations~\cite{gu2023mamba,yang2024gla,dao2024transformers}. This changes what the runtime treats as history: instead of appending independent cache slots, the model carries forward a compact state that every new token mutates.

To make this execution model concrete, we use a minimal recurrent update and omit normalization, query activations, and output projection:
\begin{equation}
  \label{eq:state-update}
  \mathbf{S}_t = \mathbf{S}_{t-1} + \mathbf{v}_t\mathbf{k}_t^\top .
\end{equation}
Here $\mathbf{S}_t$ summarizes the prefix through projected factors $\mathbf{k}_t$ and $\mathbf{v}_t$; adding a token mutates this state instead of appending an independently addressable cache slot. This stateful line of work progresses from Mamba's selective SSM backbone~\cite{gu2023mamba}, to gated matrix-state linear attention in GLA~\cite{yang2024gla}, the SSM-attention duality of Mamba2~\cite{dao2024transformers}, and DeltaNet's retrieval-oriented delta rule~\cite{yang2024deltarule}. GDN combines gating with delta-rule updates and improves language modeling, reasoning, retrieval, length extrapolation, and long-context performance~\cite{yang2025gdn}; it is also used in practical hybrid models such as Qwen3-Next, which reports higher long-context throughput than Qwen3-32B-Base~\cite{qwen2025qwen3nextcard}, and the official implementation reports integration into Qwen3-Next and Qwen3.5~\cite{nvlabs2026gdnrepo}. We therefore focus on GDN as the target layer and treat it as a recurrent-state layer whose dense persistent state is updated by projected token factors.

\stitle{Existing linear-attention kernels.} Existing kernels for linear attention can be classified into two categories as follows:

\emph{Long-sequence prefill kernels.} Existing prefill kernels for linear attention and gated variants such as GLA parallelize long, regular sequences by tiling or chunking the input, summarizing local recurrent updates, and combining block summaries with scan-style parallelization~\cite{yang2024fla,qin2024lightning,beck2025tfla}. Delta-rule variants such as DeltaNet and Gated DeltaNet introduce stronger state dependencies, so long-sequence kernels use the UT transform to rewrite token-level dependencies into block-local lower-triangular solves before merging the result into the recurrent state~\cite{yang2024deltarule,yang2025gdn}. These prefill kernels are effective when hundreds or thousands of tokens amortize layout conversion, synchronization, launch overheads, and intermediate tensors.

\emph{Single-token decode kernels.} Autoregressive decode uses the opposite execution shape: a fused recurrent kernel advances the state one token at a time and keeps the per-step computation close to the recurrence itself~\cite{yang2024fla}. This path avoids long-sequence block scans, layout conversion, and intermediate summaries. However, each step still reads the previous recurrent state and writes the updated state back at the token boundary, so the fixed state traffic remains exposed.

\stitle{Decode remains memory-bound.} Removing the growing KV cache reduces the asymptotic memory footprint, but it does not make decode compute-bound. Stateful linear-attention models keep a dense recurrent state that must be read before each token update and written back afterward. For key and value dimensions $d_k$ and $d_v$, a GDN-style state has $\Theta(d_kd_v)$ elements per value head; each token performs $\Theta(d_kd_v)$ arithmetic while moving the same dense state through memory. With $h_v$ value heads and element width $w$, the persistent state occupies $h_vd_kd_vw$ bytes, and each decode step moves $\Theta(h_vd_kd_vw)$ bytes in each direction. For the Qwen3-Next-style configuration studied by Gupta et al.~\cite{gupta2026persistent} ($h_v{=}32$, $d_k{=}d_v{=}128$, FP32), the recurrent state is 2\,MiB, and prior work characterizes batch-1 GDN decode as having arithmetic intensity below 1\,FLOP/B. Thus, the bottleneck shifts from a growing KV cache to repeated movement of a fixed but large recurrent state. This memory-bound decode path also creates an opportunity: if speculative decoding can verify multiple draft tokens in one target-side invocation, fixed launch and state-movement costs can be shared across accepted tokens. We therefore explore combining speculative decoding with linear-attention targets.

\subsection{Naive Solutions of Speculative Decoding with Linear Attention}

Given this opportunity, a natural question is whether speculative decoding can be obtained by simply plugging existing linear-attention kernels into a Transformer-style runtime. The answer is no: speculative verification creates short chain- or tree-shaped candidate workloads, while existing kernels are optimized for either single-token decode or long regular prefill.

\stitle{Naive adaptation 1: replay candidates with decode kernels.}
The first baseline treats speculative verification as a sequence of ordinary recurrent decode steps. For a chain proposal, the target applies the single-token GDN decode kernel to each draft token in order and then runs posterior selection. This preserves the original recurrence, but it also preserves the main bottleneck: each candidate step reads the dense recurrent state from memory and writes the updated state back before the next candidate can run. For tree proposals, different branches require different intermediate states, so the runtime must either snapshot full recurrent states, replay shared prefixes, or serialize branch verification through a shared state object. \emph{Problem: decode-kernel replay keeps the per-token recurrent-state round trip, while tree proposals further enlarge this cost.}

\stitle{Naive adaptation 2: use prefill kernels on draft chains.}
The second baseline starts from the opposite end and treats the submitted draft as a short prefill sequence. For a chain proposal, this means packing the draft tokens as a short causal sequence and invoking an existing parallel linear-attention or GDN prefill kernel. This exposes token-level parallelism, but the speculative window is too short to amortize the setup costs that make prefill kernels efficient, including layout conversion, block summaries, scan or UT-transformed solves, synchronization, and intermediate tensors. A tree proposal further breaks the regular sequence assumption: ordinary causal prefill lets every earlier candidate affect every later one, while each candidate should only see updates along its own prefix path. \emph{Problem: prefill-kernel reuse pays long-prefill setup costs on short chain proposals and its regular causal dependency model does not match tree proposals.}

These failures expose an abstraction mismatch rather than a single kernel inefficiency. Prior SSM and hybrid speculation work reaches a similar lesson~\cite{wu2024snakes,wu2025stree}; GDN further requires the Delta-rule factorization developed in Section~\ref{sec:parallel-factorized}.

\subsection{Challenges}

The memory-bound decode path motivates speculation, but the naive adaptations above show that batching draft tokens is not enough. Combining speculative decoding with stateful linear-attention targets raises three challenges across target verification, accepted-state recovery, and draft construction.

\stitle{Challenge 1: speculative verification struggles to reuse recurrent state in SRAM.} Linear-attention decode is efficient only if recurrent-state movement is amortized, but speculative verification gives the target a short candidate workload rather than a long regular sequence. Reusing single-token decode kernels writes the dense state back between adjacent candidates, while invoking long-sequence kernels on a small window pays layout conversion, scan or UT-solve, synchronization, and intermediate-tensor overheads that were meant to be amortized over much longer inputs~\cite{yang2024fla,beck2025tfla}. Branching drafts add another mismatch: candidate dependencies follow proposal topology rather than a regular causal sequence. The target operator therefore needs verification paths that keep state tiles resident when the workload is chain-like, preserve ancestor-only dependencies when it branches, and avoid prefill-style setup costs when the window is short.

\stitle{Challenge 2: speculative acceptance lacks cheap rollback for dense recurrent state.} During verification, each candidate can produce state-update factors before posterior selection knows whether that candidate lies on the accepted path. Unlike Transformer KV entries, these updates cannot be committed speculatively and later removed by truncating a suffix. Full-state snapshots provide rollback but move dense endpoint states, while token replay avoids snapshots but reruns factor generation that verification already performed. Prior stateful-speculation work identifies the same need for explicit state advancement~\cite{wu2024snakes,wu2025stree}. The challenge is to recover the accepted recurrent state from compact verification artifacts and commit it without adding another full recurrent-state access pass.

\stitle{Challenge 3: speculative drafting must adapt to recurrent target dynamics and costs.} Even with efficient target-side mechanisms, speculation helps only if the submitted candidates are likely to be useful. Existing draft heads and tree-construction policies are largely designed around Transformer hidden states and KV-cache verification~\cite{li2024eagle,li2024eagle2,cai2024medusa}; their scores and features may not match a linear-attention target's recurrent execution path. Draft construction also has a different cost profile: low-confidence long paths or broad branches waste state-dependent verification work even though final acceptance remains target-side. The runtime therefore needs target-aligned proposal scores, confidence-guided pruning, and a topology representation that the verifier can consume directly.

Together, these challenges require shifting the runtime abstraction from KV-cache suffix manipulation to the three-layer design of \ours: topology-aware verification, accepted-factor state management, and target-aligned draft construction.

\section{System Overview}
\label{sec:overview}

\ours follows the three stages of speculative decoding introduced in Section~\ref{sec:background}, but optimizes them in a bottom-up order. Level~1 corresponds to the target-verification stage: the target kernel must first make verification over short chain or branching candidate workloads efficient, otherwise speculative decoding cannot amortize recurrent-state movement. Level~2 corresponds to the acceptance and state-update stage: once verification produces candidate factors, the runtime must commit only accepted progress without polluting the durable recurrent state. Level~3 corresponds to the draft-construction stage: after the target-side cost and state semantics are fixed, the drafter is adapted to prune and pack candidate work that those lower-level mechanisms can use effectively.

\begin{figure}[t]
  \centering
  \includegraphics[width=\columnwidth]{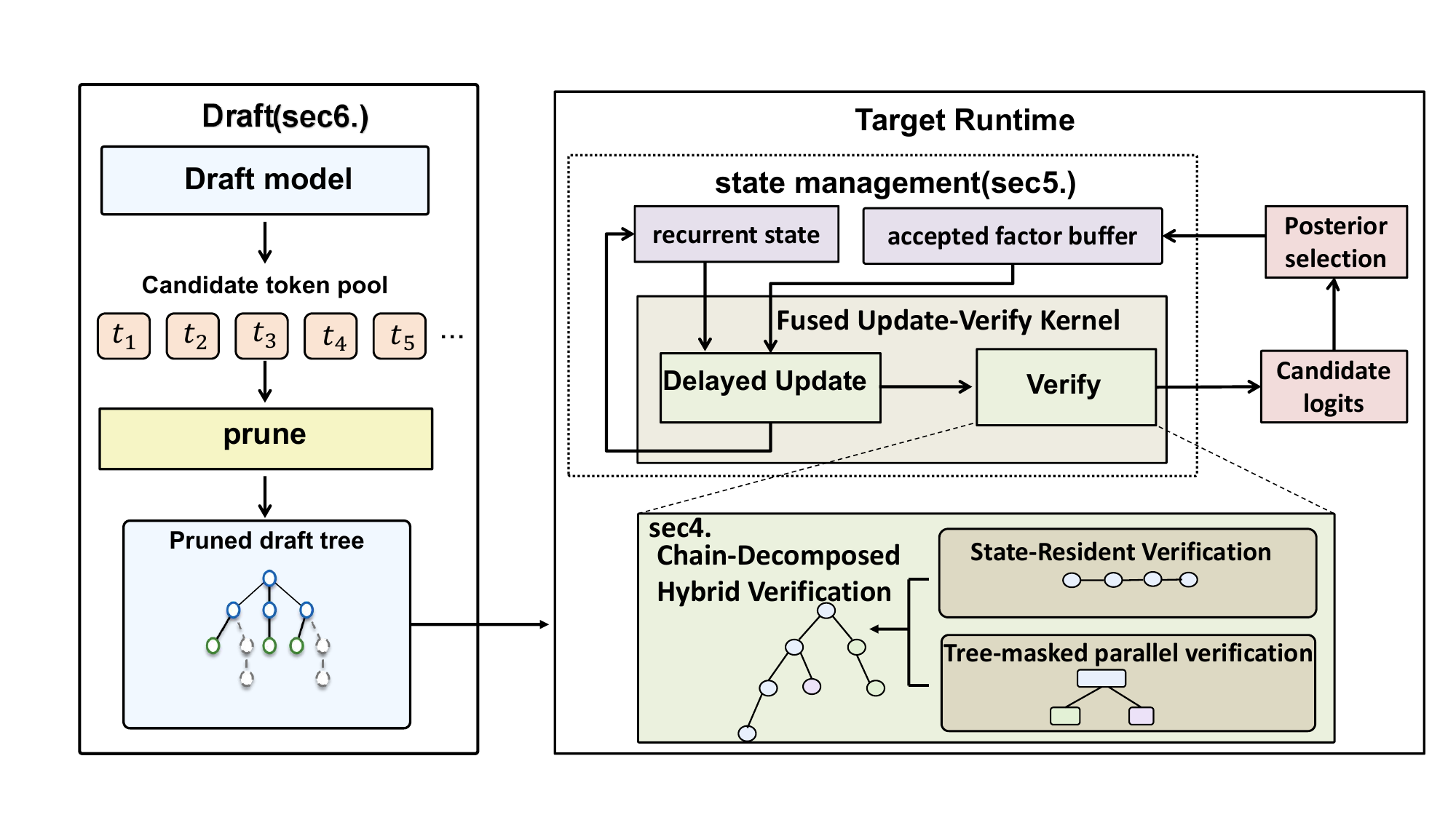}
  \caption{Overview of \ours.}
  \label{fig:overview}
\end{figure}

\stitle{Level 1: topology-aware target verification.}
Given a committed recurrent state and a packed draft topology, the verifier cannot assume that candidates form a regular causal sequence. Section~\ref{sec:kernel} specializes execution to the candidate shape: a state-resident serial path keeps state tiles live across adjacent same-layer updates for short chains; a tree-masked factorized path inserts the tree dependency mask into the GDN UT transform for branching drafts; and a chain-decomposed hybrid route runs serial kernels within dependency-respecting chains while exposing parallelism across ready chains.

\stitle{Level 2: speculative state lifecycle.}
Unlike a Transformer KV cache, a recurrent linear-attention state cannot be rolled back by truncating cache slots. Section~\ref{sec:state-management} keeps speculative factors separate from durable state: instead of snapshotting full candidate states or replaying accepted tokens, the runtime buffers the factors produced during verification and gathers only the accepted path after posterior selection. The accepted factors are then kept as a pending buffer and applied inside the next verification kernel, avoiding a standalone state-update pass.

\stitle{Level 3: integration with the speculative decoding stack.}
Efficient target-side mechanisms only help when the submitted candidates are likely to be useful. Section~\ref{sec:layer-aware-stack} prunes low-probability proposal paths before verification, reducing wasted recurrent work while leaving final acceptance to the target. It also trains an EAGLE-style drafter on recurrent features from the linear-attention target, then packs the retained candidates as parent pointers consumed by the topology-aware verifier.

\section{Topology-Aware Verification Kernels}
\label{sec:kernel}

As discussed in Section~\ref{sec:background}, existing linear-attention kernels mainly target two cases: long-sequence prefill and single-token decode. In this section, we first propose two optimized kernels: 1) a state-resident serial kernel optimized for memory access in verification; 2) a tree-masked parallel kernel that fully resolves tree dependencies of prefill kernels to support tree-structured verification. Observing that the serial kernel suffers from limited token-level parallelism while the parallel kernel suffers from additional reduction overhead, we further propose a chain-decomposed hybrid verification kernel that decomposes the draft tree into dependency-respecting chains, verifies each chain with the state-resident serial kernel, and executes chains in parallel once their boundary states are ready.

\subsection{State-Resident Serial Verification}
\label{sec:serial-explicit}

\begin{figure}[t]
\centering
\includegraphics[width=\columnwidth]{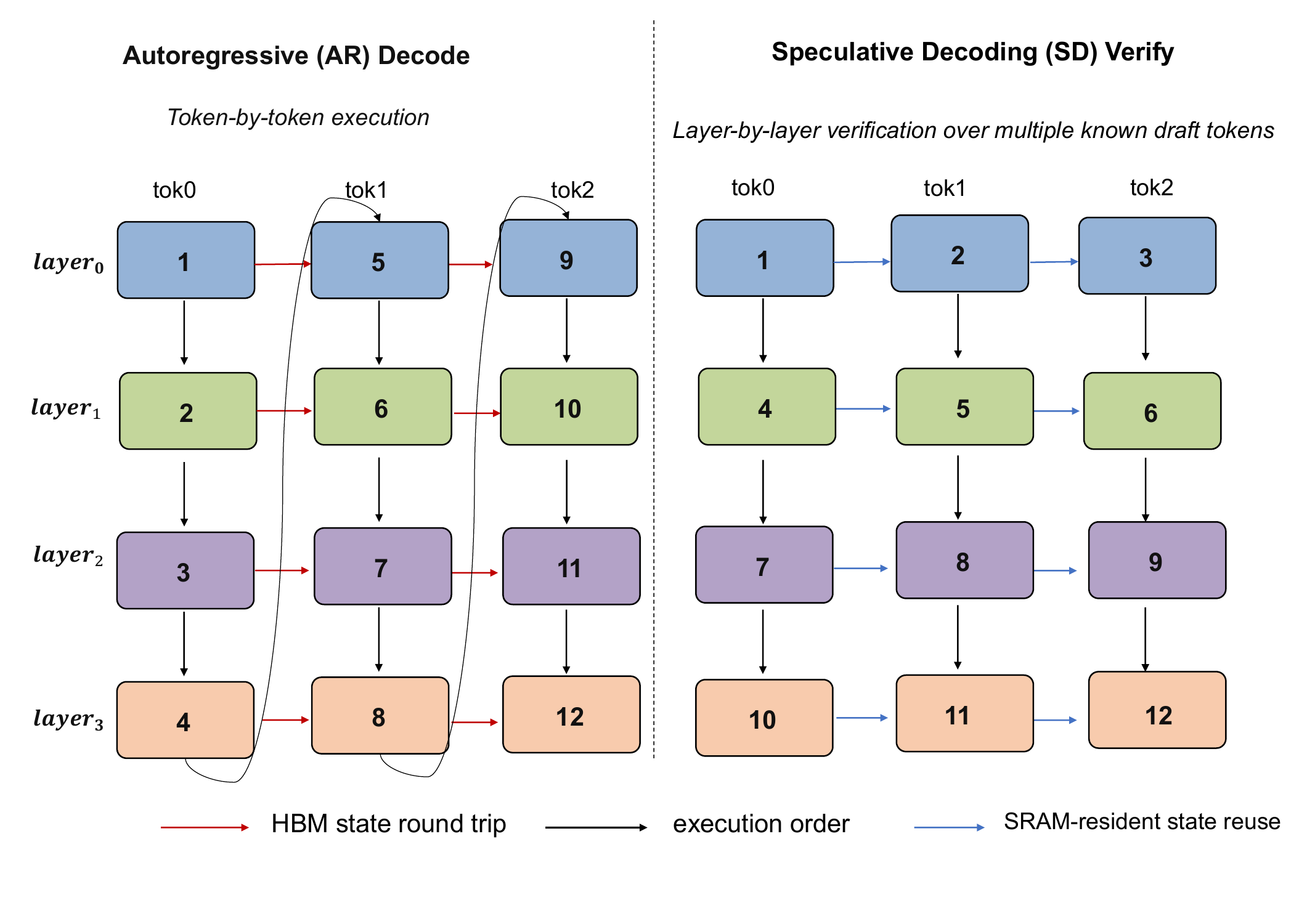}
\caption{Execution order in autoregressive decoding and speculative verification.}
\label{fig:ar-vs-sd-execution-order}
\end{figure}

\stitle{Employing decoding kernels for verification.} 
As shown in Figure~\ref{fig:ar-vs-sd-execution-order}(a), directly applying autoregressive decoding kernels will repeatedly forward each token through all layers, while the state is written back to HBM after each token and read again for the next token. This creates a large state round-trip cost between the GPU and HBM, which is especially expensive as the state size is up to 2\,MiB per layer, resulting in a 4\,MiB read--write round trip per token. 

To avoid this cost, we reorganize the execution order from token-major to layer-major as shown in Figure~\ref{fig:ar-vs-sd-execution-order}(b). In speculative verification, the draft tokens are already known. The target can verify them layer by layer, so adjacent tokens at the same layer execute consecutively. 
This schedule removes the layer gap between neighboring token updates, provides an opportunity to keep the recurrent state resident on SRAM across candidate tokens, and therefore avoids the round-trip cost of writing the state back to HBM after each token. 

However, the state size is still too large to fit on SRAM of each SM: the GDN-1.3B recurrent state is 2\,MiB per layer, while an H100 SM has only 228\,KiB SRAM. We therefore tile the state and use V-dimension tiling in the serial verification kernel.

\begin{figure}[t]
\centering
\includegraphics[width=\columnwidth]{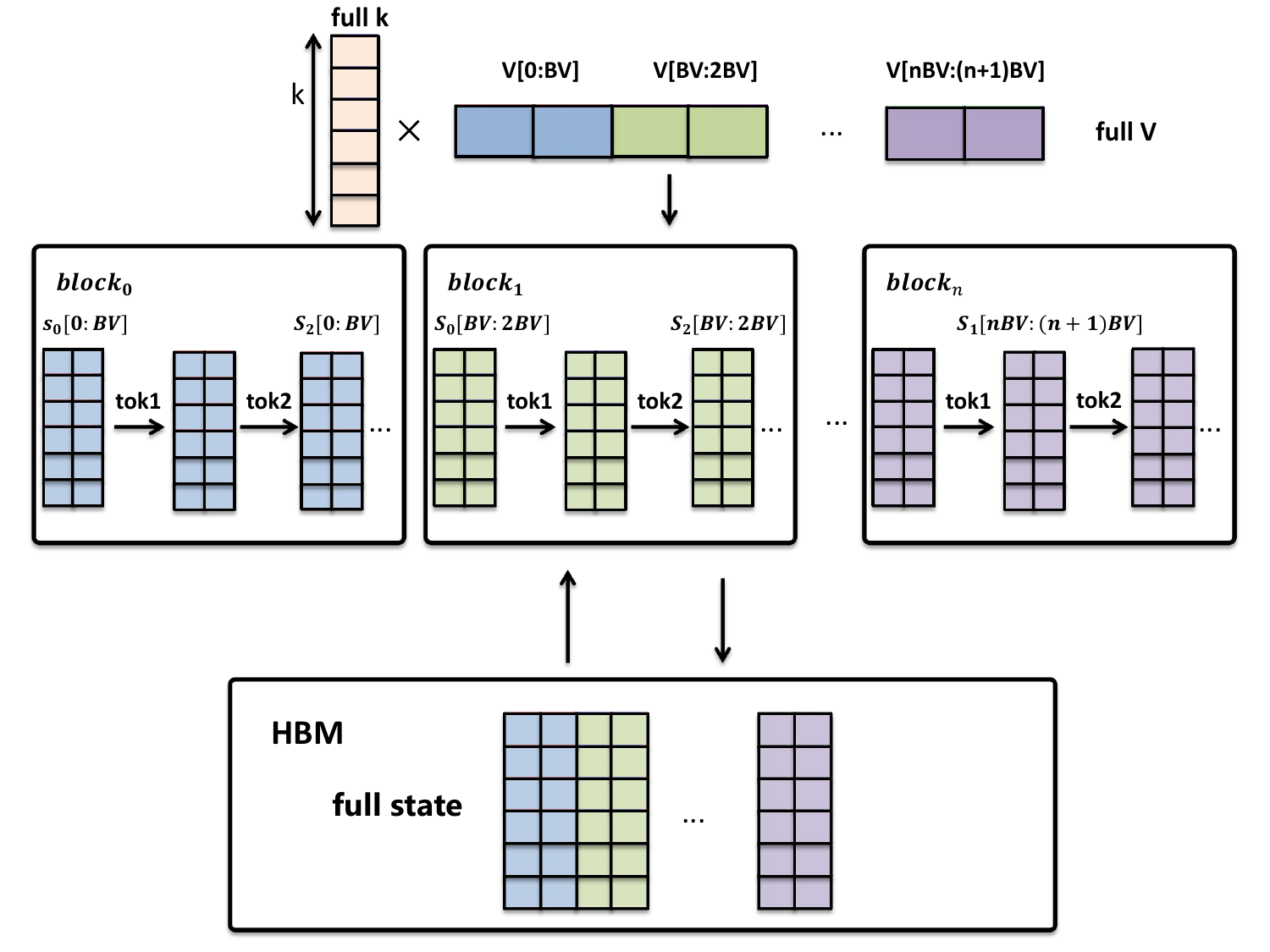}
\caption{Serial chain verification with V-dimension tiling.}
\label{fig:chain-serial}
\end{figure}

\stitle{Tiling.} The linear-attention update in Equation~\eqref{eq:state-update} adds an outer product $\mathbf{v}_t\mathbf{k}_t^\top$ to the recurrent state, so the state can be tiled along either the key or value dimension. K-dimension tiling reduces the local state footprint but partitions $\mathbf{o}_j=\mathbf{q}_j^{\top}\mathbf{S}_j$ across blocks, requiring a cross-block reduction for every token. We instead tile along the value dimension $V$ while keeping the full $K$ dimension inside each block, as shown in Figure~\ref{fig:chain-serial}. Each block updates a $K{\times}V_{\text{tile}}$ slice, computes the corresponding output slice locally, and writes back the updated slice, avoiding cross-block reductions while preserving the serial token recurrence.

\subsection{Tree-Masked Parallel Verification}
\label{sec:parallel-factorized}

The state-resident serial kernel is a low-overhead path for short chains, but it still verifies candidate tokens in dependency order and therefore cannot exploit the branch parallelism in tree-shaped drafts. The opposite prefill-style path exposes token-level parallelism, but a submitted tree is not a regular causal sequence: flattening it into root-to-leaf chains duplicates shared-prefix work, while a standard causal mask would incorrectly let sibling branches influence each other. We therefore build a fully factorized tree verifier directly around the draft topology.

\begin{figure}[t]
\centering
\includegraphics[width=\columnwidth,trim=0 24bp 0 140bp,clip]{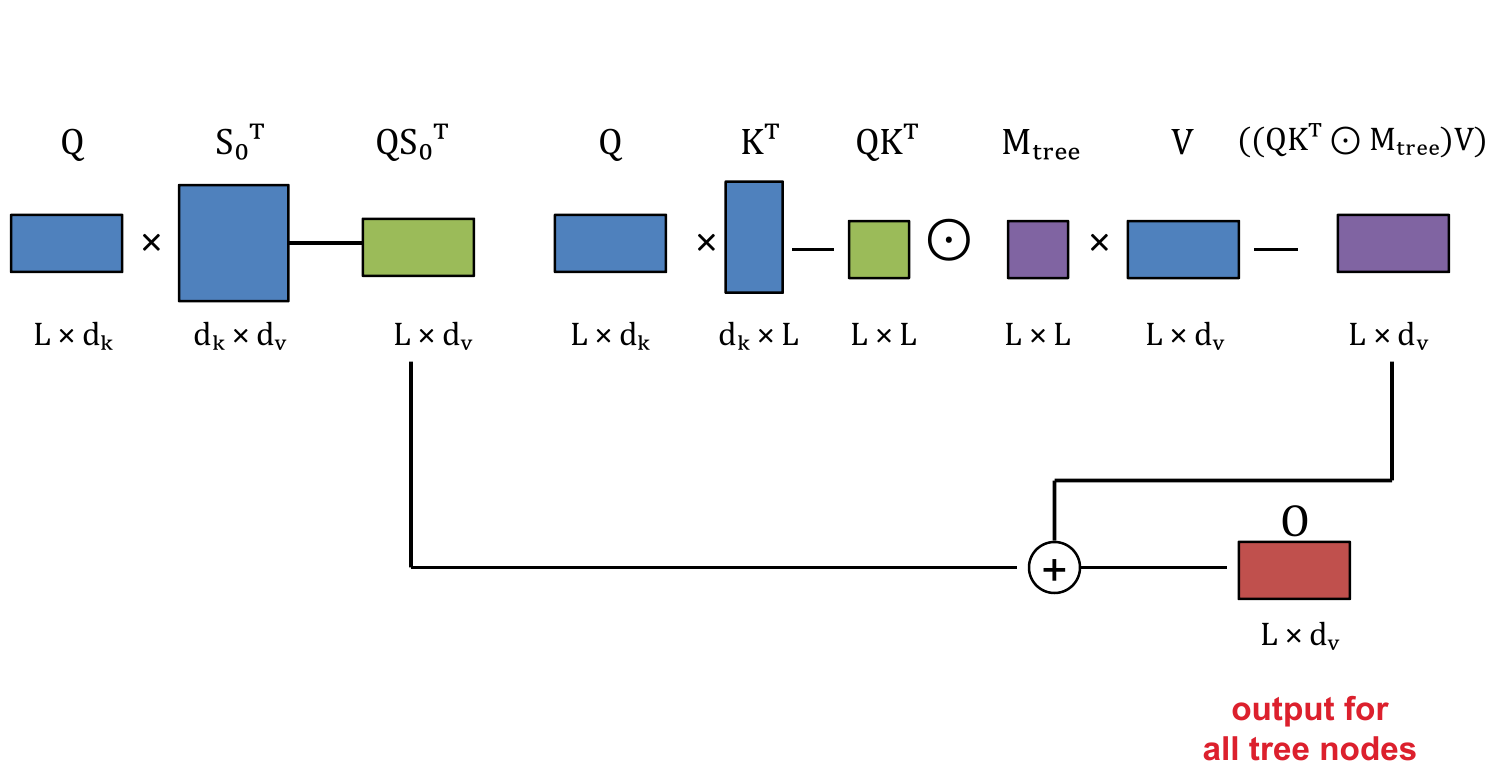}
\caption{Tree-masked parallel verification.}
\label{fig:tree-masked-parallel}
\end{figure}

Given a submitted draft tree with $L$ candidate nodes, we follow tree-based speculative verification~\cite{miao2024specinfer,wu2025stree}: the nodes are ordered topologically, and the tree mask $\mathbf{M}_{\mathrm{tree}}\in\{0,1\}^{L\times L}$ encodes ancestor-descendant state-flow dependencies while preventing sibling leakage across branches. For a standard linear-attention update such as Equation~\eqref{eq:state-update}, this mask is sufficient to compute the outputs $\mathbf{o}_j$ for all $L$ nodes using the tree-masked parallel form shown in Figure~\ref{fig:tree-masked-parallel}.

For GDN, the tree mask is inserted inside the Delta-rule UT factorization, because each corrected candidate update recursively depends on earlier corrected updates and the previous recurrent state~\cite{yang2024deltarule,yang2025gdn}. The verifier constructs the GDN token factors $\mathbf{K}_a$, $\mathbf{K}_b$, $\mathbf{U}_{\!src}$, and $\widehat{\mathbf{Q}}$, and replaces the regular sequence dependency with the tree-local transform
\[
  \mathbf{A}_{\mathrm{tree}} =
  \mathbf{M}_{\mathrm{tree}} \odot
  (\mathbf{K}_b\mathbf{K}_a^{\top}),
  \qquad
  \mathbf{T}_{\mathrm{tree}} =
  \operatorname{solve\_tril}(\mathbf{I}-\mathbf{A}_{\mathrm{tree}}).
\]
It then derives the compact factors $\mathbf{W}$, $\mathbf{U}$, and $\mathbf{Attn}$ from $\mathbf{T}_{\mathrm{tree}}$ and computes
\[
  \mathbf{O}_{\mathrm{GDN}} =
  \widehat{\mathbf{Q}}\mathbf{S}_0^{\top}
  -
  \mathbf{Attn}\left(\mathbf{W}\mathbf{S}_0^{\top}\right)
  +
  \mathbf{Attn}\mathbf{U}.
\]
Here $\mathbf{S}_0$ is the committed recurrent state; the first two terms read and correct its shared contribution, and the last term aggregates corrected candidate-window updates. Since $\mathbf{T}_{\mathrm{tree}}$ is built from $\mathbf{M}_{\mathrm{tree}}$, all candidate logits are produced from the same committed state while updates propagate only along ancestor paths, avoiding root-to-leaf replay and full recurrent-state snapshots. After posterior selection, Section~\ref{sec:state-management} gathers only the factors on the accepted path.

The fully factorized tree verifier is therefore the clean parallel endpoint of topology-aware verification: it preserves the exact tree dependencies, avoids root-to-leaf replay, and exposes parallel work across the submitted nodes. Its cost structure, however, is closer to short prefill than to recurrent decode. Constructing $\mathbf{A}_{\mathrm{tree}}$, solving for $\mathbf{T}_{\mathrm{tree}}$, deriving $\mathbf{W}$, $\mathbf{U}$, and $\mathbf{Attn}$, and moving these intermediate tensors require multiple stages that are hard to amortize over the short speculative windows used at runtime. The next subsection therefore introduces chain-decomposed hybrid verification, which preserves proposal topology but executes dependency-respecting chains with the lower-overhead state-resident serial kernel.

\subsection{Chain-Decomposed Hybrid Verification}
\label{sec:branch-parallel}

The serial and fully factorized kernels optimize opposite ends of the verification space. The serial kernel in Section~\ref{sec:serial-explicit} has low setup overhead and keeps a state tile live on chip, but it exposes little token-level parallelism. The tree-masked parallel kernel in Section~\ref{sec:parallel-factorized} exposes token-dimension parallelism, but constructing $\mathbf{T}$, $\mathbf{W}$, $\mathbf{U}$, and $\mathbf{Attn}$ requires separate kernel stages and intermediate reads and writes. For speculative windows, where the number of tokens is much smaller than in prefill, this setup cost and the reduction overhead from $K$-dimension partitioning are difficult to amortize.

We therefore use a chain-decomposed hybrid schedule: the submitted draft topology is decomposed into dependency-respecting chains, each chain is verified by the state-resident serial kernel, and chains whose boundary states are ready execute in parallel. We construct the chain decomposition with a heavy-light-decomposition policy, which partitions the submitted topology into node-disjoint chains that together cover all candidate nodes.

\begin{figure}[t]
\centering
\includegraphics[width=\columnwidth,trim=4bp 120bp 0 95bp,clip]{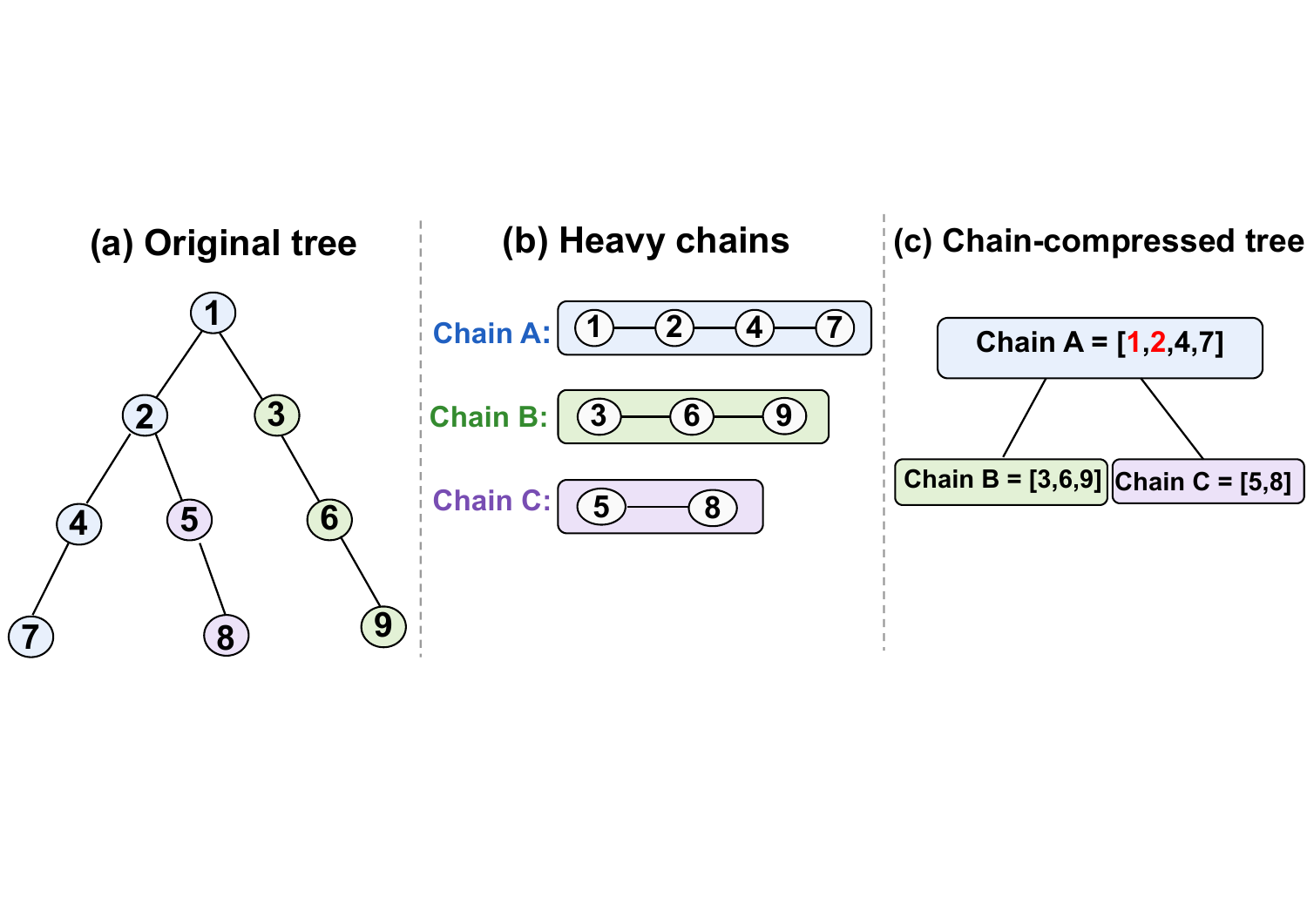}
\caption{Hybrid tree verification schedule.}
\label{fig:hld-tree-verification}
\end{figure}

Figure~\ref{fig:hld-tree-verification} shows the scheduling unit used by this design. The submitted topology in Figure~\ref{fig:hld-tree-verification}(a) is decomposed into three chains in Figure~\ref{fig:hld-tree-verification}(b): Chain A contains nodes $[1,2,4,7]$, Chain B contains nodes $[3,6,9]$, and Chain C contains node $[5,8]$. Figure~\ref{fig:hld-tree-verification}(c) then compresses each chain into one scheduling node. Chain A is the parent chain, while Chain B and Chain C start from boundary states produced inside Chain A.

For each model layer, execution follows the chain-compressed schedule in Figure~\ref{fig:hld-tree-verification}(c). Chain A is verified by the state-resident serial kernel from Section~\ref{sec:serial-explicit}; when the kernel reaches the red boundary nodes, it writes their updated recurrent states back because those states are consumed by Chain B and Chain C. Non-boundary nodes within the same chain continue to use the V-tiled state-resident execution, so their intermediate states do not need to be materialized between adjacent nodes.

The hybrid schedule is serial across model layers and parallel within a model layer. That is, layer $\ell+1$ starts only after the chain-decomposed schedule for layer $\ell$ finishes, preserving the normal layer dependency. Inside layer $\ell$, however, chains whose boundary states are ready can execute in parallel. In the example, Chain B and Chain C can be launched concurrently after the corresponding red boundary states from Chain A have been produced. This schedule preserves shared prefixes exactly once, avoids the full UT setup of the tree-masked parallel path, and still exposes parallelism across independent chains.

\section{Speculative State Management}
\label{sec:state-management}

\begin{figure*}[!t]
  \centering
  \includegraphics[width=0.99\textwidth]{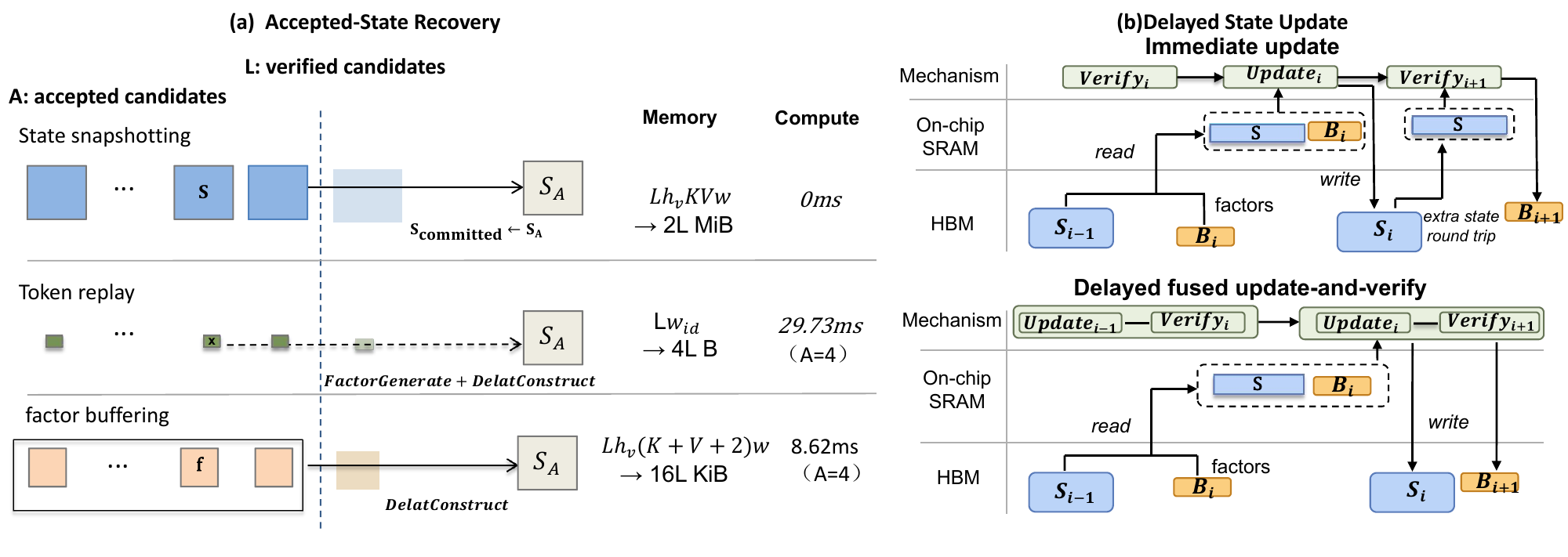}
  \caption{State management in \ours. The left side compares full-state buffering, factor buffering, and token replay at the draft-verification boundary. The right side shows how delayed state update applies the accepted-factor buffer inside the next verification kernel, reusing the same state tiles loaded from HBM and avoiding a standalone recurrent-state round trip.}
  \label{fig:state-mgmt-design}
\end{figure*}

Linear-attention targets need explicit state management for draft candidates. 
The verifier evaluates draft continuations before knowing which prefix will be accepted, yet only that accepted prefix should advance the durable recurrent state. Transformer runtimes can remove rejected draft tokens by truncating token-indexed KV-cache slots. A recurrent linear-attention state has no such suffix boundary: each candidate token updates the same dense state object, so writing draft updates into the committed state holder would let rejected candidates pollute durable model state.

In this section, we first validate two straightforward methods to recover state for accepted draft tokens: full-state snapshotting, which suffers from excessive memory consumption, and token replay, which suffers from heavy redundant computation. We then leverage the intermediate factorization of linear-attention state updates, namely factors, to trade off between memory and computation. Finally, we propose delayed state update, which applies the accepted-factor buffer at the next verification iteration rather than immediately after selection to reduce cross-kernel recurrent-state round trips.

\subsection{State Recovery for Accepted Drafts}

Before delving into our factor-buffering design, we consider two straightforward solutions to the state-recovery problem: state snapshotting and token replay. 

\stitle{State snapshotting} maintains a full state snapshot for each candidate token. As discussed in Section~\ref{sec:kernel}, the recurrent state of a linear-attention layer is a large dense tensor, so snapshotting $L$ candidates requires $L$ times the state memory. For GDN-1.3B, this is 16\,MiB per layer when $L{=}8$, which overflows the SRAM capacity of the GPU and needs to be offloaded to HBM. Despite the large memory cost, snapshotting facilitates efficient recovery: after posterior selection, the runtime copies the accepted candidate's endpoint state from HBM into the committed state holder in SRAM. However, the large memory cost and costly state copy make this approach impractical because it preserves one dense endpoint per candidate, exposing the same per-candidate state footprint that speculative decoding is meant to amortize.

\stitle{Token replay} stores only the token of each candidate and replays the accepted token sequence through the target after selection. Specifically, for each accepted token, we compute the updated state, i.e., delta state, by running the target forward on that token. Subsequently, the delta state is applied to the state to update it. Formally, we have
\begin{equation}
  \label{eq:state-update-detail}
  \begin{aligned}
  \mathbf{S}_A &= \mathbf{S}_0 + \sum_{t=1}^{A} \Delta\mathbf{S}_t,\\
  \Delta\mathbf{S}_t &= \mathrm{DeltaConstruct}\left(\mathrm{FactorGenerate}(x_t)\right).
  \end{aligned}
\end{equation}
$\mathbf{S}_0$ is the initial state before accepting any tokens, and $\mathbf{S}_A$ is the final state after accepting $A$ tokens. $\Delta\mathbf{S}_t$ is the state update contributed by the accepted token $x_t$, which can be constructed from two steps: 1) $\mathrm{FactorGenerate}(x_t)$ produces the intermediate factors for token $x_t$, which can be obtained by running the target forward on $x_t$ in the first layer or on output of previous layers; 2) $\mathrm{DeltaConstruct}$ maps these factors to the state update $\Delta\mathbf{S}_t$. However, token replay requires redundant forward computation for the accepted tokens, which is costly and negates the efficiency gain of speculative decoding. 

\stitle{Our approach: factor buffering.} As illustrated in Equation~\eqref{eq:state-update}, the state update from accepted tokens can be decomposed into two steps: factor generation and delta construction. As illustrated in Figure~\ref{fig:state-mgmt-design}(a), the $\mathrm{FactorGenerate}$ step dominates the overall computation of the state update: with a four-token accepted prefix, token replay costs 29.73\,ms, while reusing buffered factors costs 8.62\,ms, giving a 3.45$\times$ reduction. The $\mathrm{DeltaConstruct}$ step is relatively cheap. This is because the factor generation step involves heavy feedforward computation and attention projection, while the delta construction step is a lightweight vector cross-product and elementwise operations.

Regarding memory cost, the intermediate factors $f_t=\mathrm{FactorGenerate}(x_t)$ only involve a few small tensors per token, including the projected key and value vectors $\mathbf{k}_t$ and $\mathbf{v}_t$, the decay factor $\alpha_t$, and the Delta-rule coefficient $\beta_t$. Compared to the full recurrent state, these factors are much smaller in size: 128.5\,KiB per layer when $L{=}8$. 

Therefore, we cache the factors for all candidate tokens during verification and store them in a factor buffer. After posterior selection, we gather the accepted factors and apply the corresponding delta constructions to advance the state as follows:
\begin{equation}
  \label{eq:factor-buffer-update}
  \mathbf{S}_A = \mathbf{S}_0 + \sum_{t=1}^{A} \mathrm{DeltaConstruct}(f_t).
\end{equation}
This factor-buffering design achieves a better trade-off between memory and computation. It avoids the large memory cost of full-state snapshotting while also eliminating the majority of redundant computation in token replay. The factor buffer is small enough to be staged on chip tile by tile, and applying the accepted factors after selection is efficient because it only involves the cheap delta construction step.

\stitle{Remark:} We notice stateful linear layers expose different factor records: GLA-style layers need key, value, and gate/decay factors but no Delta-rule $\beta$~\cite{yang2024gla}; DeltaNet needs key, value, and $\beta$ but no GDN-style decay $\alpha$~\cite{yang2024deltarule}; Mamba-style SSMs use selective state-space parameters rather than a Delta-rule $\beta$~\cite{gu2023mamba}. Despite these differences, factor buffering remains effective because it preserves the layer-specific quantities consumed by state advancement; for each layer type, storing the relevant factors is sufficient to reconstruct accepted updates without full-state snapshots or token replay.

\stitle{Comparison of three approaches.}
Figure~\ref{fig:state-mgmt-design}(a) summarizes the trade-off. Full-state buffering minimizes post-selection computation, but it materializes one dense endpoint state per verified candidate. Token replay has the smallest speculative storage cost, but it reintroduces the target-side factor generation that verification has already performed. Factor buffering occupies the middle point: it stores more than token identifiers, but far less than full states, and turns accepted-state recovery into a lightweight delta-construction pass over the accepted records. For the GDN-1.3B dimensions in the figure, factor buffering uses ${\approx}16L$\,KiB per layer, about $128{\times}$ smaller than full-state buffering's $2L$\,MiB, while avoiding replay's repeated factor generation.

\subsection{Delayed State Update}

\stitle{Immediate update.}
A straightforward implementation updates the durable recurrent state immediately after posterior selection. Once the accepted sequence is known, the runtime gathers the accepted factors, launches a standalone state-update kernel, applies those factors to produce the base state for the next iteration, and writes the updated state back to HBM. The next verification iteration then reads this updated state again as its base state. 
As shown in Figure~\ref{fig:state-mgmt-design}(b), the standalone state-update kernel creates an extra state-access pass at every iteration boundary: the state is read and written once for the immediate update, then read again for the following verification.

\stitle{Our approach: delayed state update.}
To eliminate the extra state-access pass, we delay the state update to the next verification iteration and fuse it with the verification kernel. In other words, instead of applying the accepted factors immediately after selection, we keep them in a pending accepted-factor buffer and apply that buffer at the beginning of the next iteration's verification. 

This kernel fusion is possible because the updated state can be directly consumed by the next verification without an intermediate write-back. Specifically, the updated state tile can be kept in on-chip SRAM, eliminating the redundant read/write round trip to HBM. We partition the state-update operation so that each thread block loads a state tile, applies the relevant accepted factors, and then maps the same tile to the following verification work.

\stitle{Execution workflow.}
The delayed state-update path proceeds in three steps: 1) each iteration receives the pending accepted-factor buffer from the previous iteration together with the current draft; 2) the fused kernel applies the buffered accepted factors to the recurrent state tile and immediately verifies the current draft from the updated tile; 3) after posterior selection, the runtime gathers the factors on the newly accepted path as the pending accepted-factor buffer for the next iteration and releases all other verification artifacts, including rejected candidate factors and intermediate outputs.

\section{Integration with the SD Stack}
\label{sec:layer-aware-stack}

Sections~\ref{sec:kernel} and~\ref{sec:state-management} define the target-side mechanisms: topology-aware verification, factor buffering, and delayed state update. This section describes the draft-side integration needed to feed those mechanisms with useful candidate work. \ours keeps the external draft--verify--accept loop unchanged, but changes how candidates are formed before verification: the runtime prunes low-confidence proposals, and the drafter is trained on target-aligned recurrent features rather than borrowed from a foreign Transformer stack.

\subsection{Confidence-Guided Draft Pruning}
\label{sec:draft-pruning}

The kernels in Section~\ref{sec:kernel} make verification efficient, but they do not make every candidate worth verifying. A poor draft can still submit long chains or broad trees that the target rejects immediately, wasting recurrent verification work. \ours therefore treats draft construction as a runtime scheduling problem: the drafter should expose enough work to amortize target execution, but stop expanding branches whose scores indicate low expected acceptance.

\stitle{Tree pruning.} For tree-shaped drafts, the runtime must retain a \emph{set} of promising paths under a node budget. \ours prunes by cumulative path log-probability. After a per-layer beam top-$k$ rollout, each candidate node $v$ carries
\[
  q(v) \;=\; \sum_{i=1}^{\mathrm{depth}(v)}
  \log p_{\mathrm{draft}}\!\left(t_i \,\big|\, \mathrm{prefix},\, t_{<i}\right),
\]
i.e., the log of the path's joint draft probability. Let $q^\star=\max_v q(v)$. With tree margin $\tau_{\mathrm{tree}}$, the proposal stage keeps nodes satisfying
\[
  q(v) \ge q^\star - \tau_{\mathrm{tree}}.
\]
Because $q$ is monotonically non-increasing along any root-to-leaf path, one threshold prunes width and depth jointly---deeper paths drift below $q^\star$ even when individual steps are confident---and the window is ancestor-closed by construction. A token budget $T$ can still truncate the score-ranked selection; the runtime then joins each retained candidate's missing root-to-node ancestors until $T$ is exhausted, so the final tree size reflects both the score window and the budget.

\begin{figure}[t]
\centering
\includegraphics[width=\columnwidth]{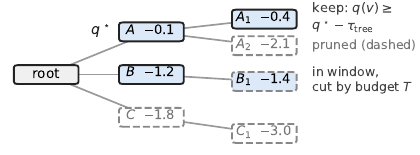}
\caption{Confidence-guided tree pruning. Nodes within $\tau_{\mathrm{tree}}$ of $q^\star$ are kept; monotonicity of $q$ makes the window ancestor-closed, and budget $T$ then truncates the selection (e.g., cutting $B_1$).}
\label{fig:draft-pruning}
\end{figure}

Figure~\ref{fig:draft-pruning} summarizes the tree-pruning policy. $\tau_{\mathrm{tree}}$ is a global window on cumulative log-probabilities that decides which paths to retain before target verification. This pruning changes only the submitted candidate set and preserves correctness because final acceptance remains target-side. The submitted tree size therefore measures the verification work that survives pruning, while accepted length measures what the target actually commits.

\subsection{Target-Aligned Drafting}
\label{sec:target-aligned-drafting}

Draft pruning reduces wasted verification only if draft scores are meaningful for the target. Pure linear-attention targets lack pretrained same-family drafters; reusing a generic Transformer drafter reduces acceptance because its hidden-state dynamics do not match the target's recurrent execution path.

\ours uses a target-aligned EAGLE-style draft layer trained on features collected from the target's own recurrent execution. During training and inference, the feature extractor observes the same recurrent-state holders used by the target decode path; the draft head learns to propose next tokens from those states. Without this alignment, the pruning signals in \S\ref{sec:draft-pruning} would be computed on features that do not reflect the target's recurrent dynamics, and margin-based scheduling would trim or retain candidates for the wrong reason. The resulting draft layer emits candidate tokens and scores, while the target runtime packs retained candidates into the chain or tree representation consumed by Section~\ref{sec:kernel}.

We deliberately keep the drafter Transformer-style rather than converting it to linear attention. The drafter proposes only a small window of tokens; at $L{\le}8$, quadratic attention is cheap---with $h{=}16$, $d{=}64$, and $L{=}8$, the KV cache is only $\approx 32$\,KiB and $O(L^2)$ work is dominated by projection and launch overhead. Making the drafter recurrent would add speculative-state management to the proposal path without exercising linear attention's asymptotic benefit at these short draft lengths. Distribution preservation and recurrent-state advancement therefore remain on the target verifier and committer.

\section{Implementation}
\label{sec:implementation}

We implement \ours in PyTorch~\cite{paszke2019pytorch} and Triton~\cite{tillet2019triton,yang2024fla}, integrated into an EAGLE-style generation runtime~\cite{li2024eagle,li2024eagle2}. The target is a single GDN layer configured to match GDN-1.3B: $h_v=8$, $d_k=d_v=256$, with a causal depthwise convolution (kernel size 4) and SiLU gating, following the GDN component in current hybrid open-weight models~\cite{qwen2025qwen3nextcard,nvlabs2026gdnrepo}. The recurrent state is stored in FP32 (2\,MiB per layer); projected activations are in BF16.

We implement four custom Triton kernels corresponding to Sections~\ref{sec:kernel} and~\ref{sec:state-management}: a serial tree-state verifier, a parallel factorized tree-state verifier, a parallel chain verifier specialized to $p_j=j-1$, and a fused commit-and-verify kernel. All kernels use Triton tile size 32 over the head dimension and process the key dimension in blocks of $B_K{=}32$; factorized kernels pad the speculative length to a multiple of 16 to control register pressure. Each kernel is validated against a sequential GDN reference advancing the recurrent state token by token.

The SD runtime is modified in four places: the cache holder stores per-layer fixed-size FP32 recurrent state $(\mathbf{S}, \mathbf{C})$ instead of KV-cache tensors; the target layer wrapper exposes a two-output forward (candidate logits and factor records $f_j$); the proposal converter transforms the drafter's tree mask into the parent array $\mathbf{p}$; and the generation loop calls \textsc{GatherAcceptedBuffer} to extract accepted factor records instead of truncating KV-cache slots.

% !TeX root = ../main.tex
\section{Evaluation}
\label{sec:evaluation}

We implement \ours in PyTorch~\cite{paszke2019pytorch} and Triton~\cite{tillet2019triton,yang2024fla}, integrated into an EAGLE-style generation runtime~\cite{li2024eagle,li2024eagle2}. The target is a single GDN layer configured to match GDN-1.3B: $h_v=8$, $d_k=d_v=256$, with a causal depthwise convolution (kernel size 4) and SiLU gating, following the GDN component in current hybrid open-weight models~\cite{qwen2025qwen3nextcard,nvlabs2026gdnrepo}. The recurrent state is stored in FP32 (2\,MiB per layer); projected activations are in BF16.

% The evaluation compares \ours against autoregressive decoding and FLA-SD, then studies end-to-end speed, tree-verification efficiency on runtime-like EAGLE proposals, accepted-state management overhead, and acceptance--draft-length regimes.

\subsection{Experimental Setup}
\label{sec:eval-setup}

\noindent\textbf{Target model and runtime.}
Our end-to-end experiments use a public GDN-1.3B checkpoint~\cite{map2026gdn13b} as the target model and an EAGLE-style GDN draft layer as the proposal model. The target is a pure Gated DeltaNet model, so the speculative runtime manages recurrent linear-attention state rather than Transformer KV-cache suffixes. The recurrent state is stored in FP32; end-to-end experiments use the correctness-first runtime path, while kernel microbenchmarks use the precision specified by the corresponding implementation path.

\noindent\textbf{Datasets and generation.}
We evaluate end-to-end generation on an EAGLE-style mixed prompt suite, GSM8K, and HumanEval. The mixed suite contains 480 prompts across conversation, instruction following, mathematical reasoning, code generation, question answering, and summarization; the GSM8K and HumanEval suites contain 1,319 and 164 prompts, respectively. All end-to-end routes use greedy decoding and generate up to 128 new tokens per prompt. The chain route uses an 8-token draft budget, while \ours uses up to 16 draft nodes with top-$k{=}4$ branching.

\noindent\textbf{Baselines and metrics.}
The end-to-end study compares \textit{Autoregressive}, which decodes one target token per step; \textit{FLA-SD}, which applies the same GDN drafter through the native FLA/GDN source path; \textit{Chain}; and \ours. The primary metric is speedup over autoregressive target decoding, computed from measured end-to-end generation latency. We also report average accepted length and first-token match, where first-token match is the fraction of speculative rounds in which the selected draft root agrees with the target greedy next token. Exact output hashes are checked against autoregressive decoding for all reported speculative routes.

\noindent\textbf{Hardware and kernels.}
All reported experiments run on an NVIDIA H100 GPU. End-to-end experiments measure the full generation call with synchronization. In the full-test data, each prompt-route pair is timed once, and dataset-level speedups are computed from summed workload latencies over matched prompts rather than from averages of per-prompt ratios. Kernel microbenchmarks repeat the same shape after warmup and report the median latency from the timing method used by the benchmark. Each custom Triton kernel is checked against a sequential or standalone GDN reference before timing.

\subsection{End-to-End Performance}
\label{sec:eval-e2e}

\begin{figure}[!htbp]
  \centering
  \includegraphics[width=0.98\columnwidth]{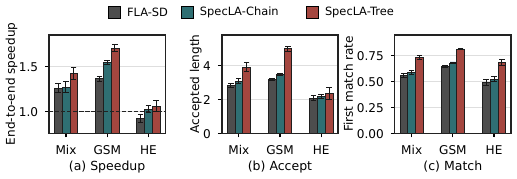}
  \caption{End-to-end performance.}
  \Description{Three horizontal bar-chart panels showing end-to-end speedup, accepted length, and first-token match for FLA-SD, Chain, and SpecLA.}
  \label{fig:eval-e2e-summary}
\end{figure}

Figure~\ref{fig:eval-e2e-summary} reports speedup, average accepted length, and first-token match for the same routes; whiskers are 95\% bootstrap confidence intervals over prompts. It shows that the optimized runtimes improve end-to-end decoding. \ours is the fastest route on all three evaluation sets, achieving 1.42$\times$, 1.70$\times$, and 1.06$\times$ speedup on the EAGLE-style mixed suite, GSM8K, and HumanEval, respectively. The chain route also improves over autoregressive decoding, reaching 1.26$\times$, 1.54$\times$, and 1.03$\times$ speedup.

FLA-SD uses the same drafter but keeps the native target-side path. It is competitive when speculative progress is sufficient, but it falls below autoregressive decoding on HumanEval.

We use \ours-Tree to denote the hybrid chain-decomposed implementation of the tree route. The speedup follows accepted speculative progress: on GSM8K, \ours-Tree increases average accepted length from 3.47 to 5.00 tokens and first-token match from 0.68 to 0.81 relative to the chain route. The mixed suite shows the same trend, with accepted length increasing from 3.08 to 3.92 and first-token match from 0.59 to 0.73. HumanEval remains the hardest workload: the tree route accepts only 2.34 tokens on average, so all speedups are smaller. These results show that target-side optimization is most useful when the drafter supplies enough accepted work to amortize recurrent verification and commitment.

\subsection{Tree Verification Efficiency}
\label{sec:eval-tree-verification}

\begin{figure}[!htbp]
  \centering
  \includegraphics[width=0.98\columnwidth]{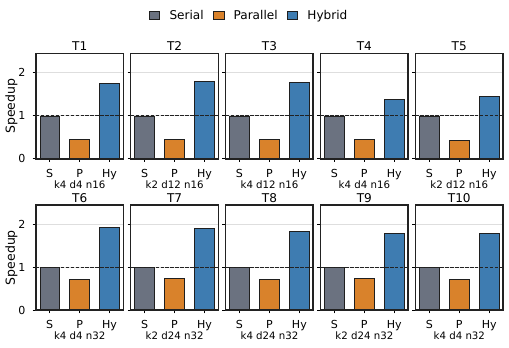}
  \caption{Tree verification examples.}
  \Description{Ten small grouped bar charts arranged in two rows. Each chart shows serial, parallel, and hybrid speedup for one selected proposal tree.}
  \label{fig:eval-tree-runtime-cases}
\end{figure}

Chain-shaped verification is evaluated with the verification-kernel design in Section~\ref{sec:kernel}. Here we focus on tree verification because it is the runtime path that differs most from ordinary recurrent decoding. Figure~\ref{fig:eval-tree-runtime-cases} reports ten example proposal trees from the recorded end-to-end benchmark at the target shape ($h_v{=}8$, $d_k{=}d_v{=}256$). We restrict this illustrative subset to runtime-like EAGLE proposals with 16--32 nodes and finite serial, parallel, and hybrid measurements, excluding pure chains and extremely wide all-root cases. Table~\ref{tab:eval-tree-shape-avg} averages over the full filtered subset.

Figure~\ref{fig:eval-tree-runtime-cases} explains the target-side cost of the measured tree route. Each mini-plot compares serial tree-state verification (S), factorized parallel verification (P), and the hybrid chain-decomposed kernel (Hy), normalized to serial verification. The labels show the selected tree id and proposal metadata $(k,d,n)$ for top-$k$, configured depth, and total nodes. The per-tree figure uses serial verification as the local baseline, whereas Table~\ref{tab:eval-tree-shape-avg} switches to root-to-leaf replay as a common cross-shape baseline. On these runtime-like trees, Hybrid improves over serial verification by 1.39--1.92$\times$, while the parallel route remains below serial and does not amortize setup cost at the actual target shape.

\begin{table}[tbp]
\small
\centering
\caption{Average tree-verification speedup by proposal shape.}
\label{tab:eval-tree-shape-avg}
\setlength{\tabcolsep}{2.8pt}
\begin{tabular}{@{}rrrrrrr@{}}
\toprule
Top-$k$ & Config. depth & Nodes & R2L (ms) & Serial & Parallel & Hybrid \\
\midrule
2 & 12 & 16 & 22.5 & 1.31$\times$ & 0.58$\times$ & 2.20$\times$ \\
2 & 24 & 32 & 55.0 & 1.36$\times$ & 1.02$\times$ & 2.55$\times$ \\
4 & 4 & 16 & 37.8 & 3.73$\times$ & 1.70$\times$ & 5.14$\times$ \\
4 & 4 & 32 & 91.2 & 4.74$\times$ & 3.47$\times$ & 7.11$\times$ \\
4 & 12 & 16 & 19.0 & 1.19$\times$ & 0.52$\times$ & 1.80$\times$ \\
4 & 24 & 32 & 41.0 & 1.30$\times$ & 0.95$\times$ & 2.19$\times$ \\
\bottomrule
\end{tabular}
\end{table}

Table~\ref{tab:eval-tree-shape-avg} aggregates the full actual-shape runtime-like subset by proposal shape using root-to-leaf replay as the common baseline. The route columns are normalized to root-to-leaf replay of the same proposal tree, and the R2L column reports that replay baseline latency in milliseconds. Serial tree-state verification already removes duplicated prefix work, while Hybrid further reduces state traffic on the branching proposal trees used by the end-to-end route. The average Hybrid speedup over root-to-leaf replay is at least 1.80$\times$ in every group and reaches 7.11$\times$ on the 32-node top-$k{=}4$ runtime-like shape.

\subsection{State-Management Overhead}
\label{sec:eval-state-mgmt}

Verification alone is not sufficient for recurrent speculative decoding. After posterior selection, only the accepted path may advance the durable linear-attention state. We evaluate two state-management choices from Section~\ref{sec:state-management}: factor-buffer reuse, which avoids replaying accepted tokens through the target body, and delayed state update through a fused commit-and-verify path, which merges accepted-state commitment with the next verification round.
In this subsection, $B$ is batch size, $h_v$ is the number of GDN value heads, and $d_k$ and $d_v$ are the key and value dimensions per head.

\begin{figure}[!htbp]
  \centering
  \includegraphics[width=0.98\columnwidth]{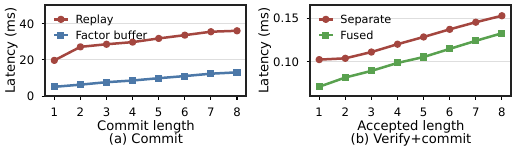}
  \caption{State-management latency.}
  \Description{Two line charts showing latency for token replay versus factor-buffer reuse and separate verify-plus-commit versus fused commit-and-verify.}
  \label{fig:eval-state-curves}
\end{figure}

Figure~\ref{fig:eval-state-curves}(a) compares packed token replay against flushing the accepted-factor buffer over 24 layers with $h_v{=}8$ and $d_k{=}d_v{=}256$. It uses the same packed replay boundary as the commit path: the accepted prefix is replayed through the target body, while factor-buffer reuse directly flushes the accepted factors produced by verification. The benchmark uses the same replay boundary for both routes and measures each commit length after warmup. Token replay median latency is 19.70 ms at commit length 1 and 35.96 ms at commit length 8, while factor-buffer reuse grows from 5.15 ms to 13.10 ms, a 2.74--4.28$\times$ median latency reduction. The committed state is unchanged in the FP32 check: the recurrent-state max difference is at most $9.54{\times}10^{-7}$ and the convolution-state difference is zero.

Figure~\ref{fig:eval-state-curves}(b) shows the delayed state-update effect with $B{=}1$, verify length 4, $h_v{=}8$, and $d_k{=}d_v{=}256$. The separate route verifies a draft from the durable base state and then runs the accepted-state commit kernel, while the fused path reads the durable state once, applies the accepted-factor buffer, verifies the current draft from the advanced state, and writes the state back. For accepted lengths 1--8, the measured separate verify-then-commit workflow grows from 0.102 ms to 0.153 ms, while fused latency grows from 0.071 ms to 0.133 ms, giving a 1.15--1.44$\times$ latency reduction. All recorded fused and standalone outputs and states match exactly in the FP32 comparison.

\subsection{Controlled Upper- and Lower-Bound Study}
\label{sec:eval-controlled}

To isolate proposal quality, we run a controlled study on the same measured 1.3B GDN target/runtime pair as the end-to-end experiments. The target, drafter, and runtime costs are measured, while token acceptance is synthetically sampled with probability $p \in \{0.5,0.6,\ldots,1.0\}$ for chain draft lengths $L \in \{5,\ldots,12\}$. The speculative route is the optimized chain runtime with draft generation, target verification, posterior acceptance, and accepted-state commitment. Because accepted tokens are forced by the assumed process, this study reports latency ratios only; correctness-preserving end-to-end results are shown in Figure~\ref{fig:eval-e2e-summary}.

We also include a calibrated 9B-scale projection to understand how the operating region shifts when the autoregressive target-only term becomes larger. This projection is not a measured 9B speculative setup. It replaces only the autoregressive target-only latency with a local Qwen3.5-9B proxy, while keeping speculative-side overhead calibrated from the measured GDN-1.3B chain runtime.

\begin{figure}[tbp]
\centering
\includegraphics[width=\columnwidth]{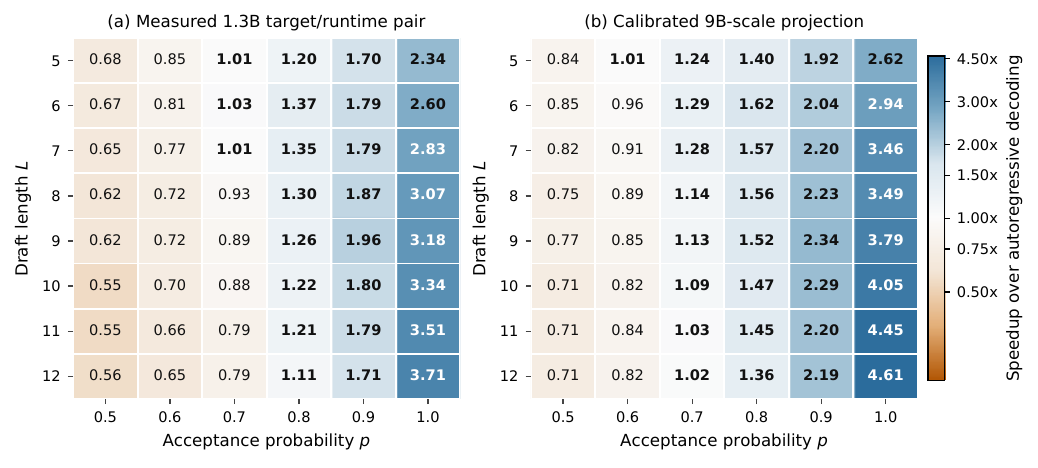}
\caption{Controlled acceptance regimes.}
\Description{Two heatmaps showing speedup as a function of acceptance probability and draft length for the measured 1.3B runtime pair and a calibrated 9B-scale projection.}
\label{fig:eval-controlled-heatmap}
\end{figure}

Figure~\ref{fig:eval-controlled-heatmap} shows that proposal quality determines the useful draft budget. On the measured 1.3B pair, $p{=}0.5$ and $p{=}0.6$ remain below 1$\times$ for every reported draft length. The transition region is around $p{=}0.7$: short drafts can slightly exceed 1$\times$, but longer drafts still lose because too few proposed tokens are committed. Once $p$ reaches 0.8, all reported draft lengths become beneficial; with oracle acceptance, speedup reaches 3.71$\times$ at $L{=}12$. The calibrated 9B-scale projection shifts the profitable region upward: all $p{=}0.7$ settings become profitable, $p{=}0.8$ gives 1.36--1.62$\times$ projected speedup, and oracle acceptance reaches 4.61$\times$. Thus, larger draft budgets raise the upper bound, but they require a sufficiently accurate drafter to pay off.

\section{Conclusion}

In this paper, we studied how to make speculative decoding effective for stateful linear-attention models, where autoregressive decoding still moves dense recurrent states one token at a time. We presented \ours, a runtime that verifies submitted chains and trees with topology-aware kernels, recovers accepted states from compact verification factors with delayed state update, and feeds the verifier with confidence-pruned candidates from a target-aligned drafter. On an NVIDIA H100 with a public GDN-1.3B target, \ours achieves up to 1.70$\times$ end-to-end speedup over autoregressive decoding.

\bibliography{reference}

\end{document}